\documentclass{article}

\usepackage{times}
\usepackage{amssymb,amsfonts}
\usepackage[fleqn]{amsmath}
\usepackage[T1]{fontenc}
\usepackage[utf8]{inputenc}

\IfFileExists{upquote.sty}{\usepackage{upquote}}{}

\IfFileExists{microtype.sty}{%
 \usepackage{microtype}
 \UseMicrotypeSet[protrusion]{basicmath} % disable protrusion for tt fonts
}{}

\usepackage[pagebackref=true]{hyperref} %sysml

\usepackage{graphicx,grffile}
\usepackage{lipsum}
\usepackage{listings}
\usepackage{afterpage}
\usepackage{endnotes,epsfig}
\usepackage{multirow}
\usepackage{xspace}
\usepackage{tabularx}
\usepackage{ragged2e}
\usepackage{booktabs}
\usepackage{paralist}
\usepackage[american]{babel}
\usepackage[shortlabels]{enumitem}
\usepackage{courier,color,wrapfig}
\usepackage{xspace}
\usepackage{balance}
\usepackage{enumitem}
\usepackage{epstopdf}
\usepackage{multirow}
\usepackage{booktabs}
\usepackage{url}
\usepackage{xurl}
\usepackage{amssymb}
\usepackage{pifont}
\usepackage{subfigure}
\usepackage{tikz}
\usepackage{comment}
\usepackage{mathtools}
\usepackage[normalem]{ulem}
\usepackage{xkeyval}
\usepackage[algo2e]{algorithm2e}
\usepackage{ifthen}
\usepackage{etoolbox}
\usepackage[margin=5pt,font=small,labelfont={bf,rm}]{caption}
\usepackage{bbding}
\usepackage{enumitem}
\usepackage{stfloats}

\usepackage{array}
\newcolumntype{L}[1]{>{\raggedright\let\newline\\\arraybackslash\hspace{0pt}}m{#1}}
\newcolumntype{C}[1]{>{\centering\let\newline\\\arraybackslash\hspace{0pt}}m{#1}}
\newcolumntype{R}[1]{>{\raggedleft\let\newline\\\arraybackslash\hspace{0pt}}m{#1}}
\usepackage{amsmath}
\usepackage[textsize=tiny]{todonotes}

\PassOptionsToPackage{usenames,dvipsnames}{color} % color is loaded by hyperref

\makeatletter
\def\maxwidth{\ifdim\Gin@nat@width>\linewidth\linewidth\else\Gin@nat@width\fi}
\def\maxheight{\ifdim\Gin@nat@height>\textheight\textheight\else\Gin@nat@height\fi}
\makeatother
\setkeys{Gin}{width=\maxwidth,height=\maxheight,keepaspectratio}


\providecommand\parab[1]{\noindent\textbf{#1}}

\apptocmd\normalsize{%
\abovedisplayskip=5pt
\abovedisplayshortskip=5pt
\belowdisplayskip=5pt
\belowdisplayshortskip=5pt
}{}{}

\hypersetup{final}
\newcommand{\sysname}{\textsc{FluidML}\xspace}

\newcommand{\etc}{\textit{etc.}\xspace}
\newcommand{\ie}{\textit{i.e.,}\xspace}
\newcommand{\eg}{\textit{e.g.,}\xspace}

\newcommand{\figref}[1]{Figure~\ref{#1}}
\newcommand{\tabref}[1]{Table~\ref{#1}}
\newcommand{\algoref}[1]{Alg.~\ref{#1}}
\newcommand{\eqnref}[1]{Eqn.~\ref{#1}}

\newcommand{\squishlist}{
 \begin{list}{$\bullet$}
 		{ \setlength{\itemsep}{0pt}
 			\setlength{\parsep}{3pt}
 			\setlength{\topsep}{3pt}
 			\setlength{\partopsep}{0pt}
 			\setlength{\leftmargin}{1.5em}
 			\setlength{\labelwidth}{1em}
 			\setlength{\labelsep}{0.5em} } }

\newcommand{\squishend}{
  \end{list}  }

\usepackage[accepted]{mlsys2025}

\mlsystitlerunning{\sysname}

\begin{document}

\twocolumn[
\mlsystitle{\sysname: Fast and Memory Efficient Inference Optimization}
% \mlsystitle{\sysname: Fast and Memory Efficient Inference Runtime}

% FuildML: fast and lean universal inference design for edge ML?
% FuildML: fast, lightweight, and universal inference design?

\begin{mlsysauthorlist}
\mlsysauthor{Jinjie Liu}{goo}
\mlsysauthor{Hang Qiu}{to}
\end{mlsysauthorlist}

\mlsysaffiliation{to}{Department of Electrical and Computer Engineering, University of California Riverside}
\mlsysaffiliation{goo}{Department of Electrical and Computer Engineering, University of Southern California}

\mlsyscorrespondingauthor{Hang Qiu}{hangq@ucr.edu}

\mlsyskeywords{Edge ML, ML Deployment Validation, ML Debugging}

\vskip 0.1in

\begin{abstract}
Machine learning models deployed on edge devices have enabled numerous exciting new applications, such as humanoid robots, AR glasses, and autonomous vehicles. However, the computing resources available on these edge devices are not catching up with the ever-growing number of parameters in these models. As the models become bigger and more complicated, the novel yet sophisticated structure challenges the inference runtime optimization. We present \sysname, a generic runtime memory management and optimization framework that can flexibly transform the model execution blueprint to achieve faster and more memory-efficient inference. Evaluations across different platforms show that \sysname can consistently reduce the end-to-end inference latency by up to 25.38\% for popular language models and reduce peak memory usage by up to 41.47\%, compared to state-of-the-art approaches. \sysname is of $\sim$30K line of codes, built for general-purpose usage, and will be released as an open-source inference runtime optimization framework to the community.
\end{abstract}

]

% \printAffiliationsAndNotice{}  % leave blank if no need to mention equal contribution
% \printAffiliationsAndNotice{\mlsysEqualContribution} % otherwise use the standard text.

\section{Introduction}
\label{sec:Intro}

Taking advantage of near-sensor inference, machine learning (ML) models deployed on edge devices enabled many low-latency, low-power, and privacy-sensitive applications. Nevertheless, the constant tension between limited resources and ever-growing model sizes (\eg transformers, foundation models) continues to hinder the deployment, which in turn keeps driving the research in optimizing edge inference latency and squeezing memory footprint. 

From an ML perspective, the state of the practice is to employ advanced \textit{model shrinking} techniques such as pruning~\cite{DBLP:journals/corr/HanMD15}, quantization~\cite{DBLP:journals/corr/HanMD15}, knowledge distillation~\cite{journals/corr/HintonVD15}, neural architecture search~\cite{DBLP:journals/corr/ZophL16}, \etc These techniques aim at less memory footprint and faster inference by identifying and eliminating redundancy in the model itself without compromising performance. From the system perspective, given an optimized/shrunk graph of connected operators, researchers have proposed \textit{kernel optimization} for fast and memory-efficient execution~\cite{234946}, including constant folding, activating fusion, and using high-performance computing libraries (\eg BLAS, OpenBLAS~\cite{6413635,6877458}), \etc These techniques either reduce the memory copy overhead or enable parallelism to cut end-to-end inference latency. 

Today, modern vision, language, and foundation models have become more sophisticated, rendering complicated interweaving graphs that connect both conventional and novel operators. These graphs are no longer a single branch of homogeneous kernels like traditional convolutional neural networks (CNNs). Hence, independently optimizing individual kernels may not yield the best performance; this approach can easily overlook larger opportunities for joint optimization. Fundamentally, there is a lack of a generic, model-agnostic framework that can provide a holistic plan for how the numeric computation should flow throughout the graph. 

In this paper, we present \sysname, a generic memory management and optimization framework that can flexibly transform the execution blueprint to achieve faster and more memory-efficient inference. Specifically, \sysname 1) jointly optimizes the operator memory layout throughout the entire graph, 2) speeds up both operator and end-to-end graph execution by producing a memory access-friendly execution blueprint, 3) reduces peak memory usage by carefully allocating resources according to the optimized runtime, 4) provides a VM to collect real-world data and helps the compiler to find the best schedule.

% \hqrev{Jinjie please check the following paragraph and correct as you see fit}\jjrev{It looks good to me.}
We implement \sysname using Open Neural Network Exchange (ONNX) as the front-end format, supporting tens of widely used operators, and built our \sysname compiler
% \footnote{Please refer to the submitted artifact. Code will be open-sourced upon acceptance.} 
on top of LLVM and Multi-Level Intermediate Representation (MLIR). \sysname is approximately 30K lines of C++ code, and we build it for general-purpose usage on different platforms.
Evaluation using the Just-in-Time (JIT) Execution Engine on three different platforms (Intel, AMD, Aarch64) shows that \sysname can consistently reduce the end-to-end inference latency by up to 25.38\% for popular language models (\eg BERT~\cite{kim2021bert}), up to 17.87\% for widely used operators (\eg MatMul), and reduce peak memory usage by up to 41.47\%.

In summary, we make the following contributions:

\begin{itemize}[leftmargin=4mm]
    \vspace{-2mm}
    \item We introduce a generic joint graph-operator optimization framework to transform the ML execution blueprint to be faster and more memory-friendly.
    \vspace{-2mm}
    \item We provide a corresponding memory management algorithm to reduce the peak inference footprint.
    \vspace{-2mm}
    \item We show that \sysname can find global optimization opportunities across different models on various platforms.
    \vspace{-6mm}
    \item We demonstrate plenty of space for inference runtime optimization, raising awareness of the impact of memory access patterns during inference.
    \vspace{-2mm}
\end{itemize}
The code will be open-sourced to the community.

\section{Background and Motivation}
\label{sec:background}

% \subsection{Ahead-of-Time (AOT) Compilation.}
In this section, we describe the current state of practice for model inference and the remaining challenges and opportunities that motivate the design and implementation of \sysname. 
There are two main approaches to implementing model inference: inference engine and compiler. To facilitate frequent iterations in the model development and testing phase, many popular inference frameworks widely adopt the inference engine, including the early versions of PyTorch~\cite{DBLP:journals/corr/abs-1912-01703}, TensorFlow~\cite{tensorflow2015-whitepaper}, ONNXRuntime~\cite{onnxruntime}, \etc The inference engine provides flexibility to handle different model architectures and tensor shapes, which allows developers to iterate faster and more conveniently. However, the architecture, parameters, and tensor shapes are frozen when the model enters the deployment phase; the numeric computation is assumed a certain pattern inherited from the model design phase without much thought given to the hardware execution. This approach limits the flexibility at runtime, which often misses latency enhancement opportunities.

\parab{Ahead-of-Time (AOT) Compilation.}
On the other hand, the Ahead-of-Time compilation could seize such opportunities to manipulate operators and graphs into an equivalent variant optimized for execution. For example, techniques including constant folding, common sub-expression elimination, and operator fusion have been widely adopted in AOT compilation (\eg TensorFlow Lite, PyTorch Mobile). Other popular deep learning compilers includes  MLIR~\cite{DBLP:journals/corr/abs-2002-11054, mlir, DBLP:journals/corr/abs-2003-00532}, TVM~\cite{DBLP:journals/corr/abs-1802-04799}, IREE~\cite{The_IREE_Authors_IREE_2019}, \etc

The key advantage of AOT compilation is its \textit{lean} and \textit{versatile} intermediate product. The inference engine often carries a tree of dependency libraries, which is a huge burden for inference deployment, especially in runtime environments where users need to install all the dependencies, running into the  "Dependency Hell" to make it work. This problem is exacerbated on resource-constrained platforms (\eg system-on-chip (SoC) or embedded systems) without operating system support. Instead, the AOT compiler could provide low-level intermediate codes like C/C++ code, LLVM intermediate representation (LLVM IR), or assembly, which can be easily converted into executable files suitable for running on these platforms.

At runtime, the compiled intermediate products can be stored in the read-only area (\eg, the ".rodata" segment for x86-64) for fast access, while the inference engine must load dynamically allocate buffered on demand and load the model parameters into these buffers. Also, modern operating systems and compilers provide more optimizations for such data types (\ie .rodata), including decreasing the memory cost and increasing the memory accessing speed.

\begin{figure}
    \centering
\subfigure[Row-Major]{
\includegraphics[width=0.8\linewidth]{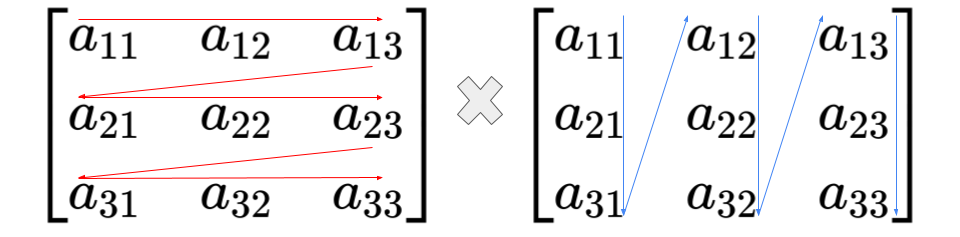}}
\subfigure[Column-Major]{    
\includegraphics[width=0.8\linewidth]{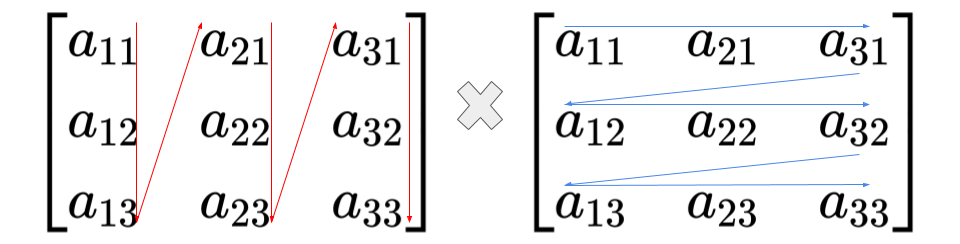}}
\centering
\caption{Memory access patterns for matrix multiplication}
\label{fig:matmul}
\end{figure}

\parab{Memory Layout in Kernel Implementation.}
There have been many studies on the impact of tensor memory layout on operator execution efficiency~\cite{DBLP:journals/corr/abs-1802-04799,DBLP:journals/corr/abs-1805-00907}. A better algorithm plays a vital role in accelerating the operator execution speed. Still, a better data access pattern could also increase the cache hit accuracy and decrease the time cost of the memory access. Take general matrix multiply (GEMM~\cite{10.5555/509058.509096,10.1007/11558958_2,10.1145/1356052.1356053}) as an example, \figref{fig:matmul} shows two different memory access patterns; row-major and column-major. For a typical GEMM operator whose left input with the shape of $(i,k)$, right input with shape $(k,j)$, and output with the shape $(i,j)$, all of the tensors are stored in row-major in the memory. The traditional implementation of the GEMM is to access these three axes in the order of $i,j,k$. However, for the right input, the order of $j,k$ conflicts with the right input's memory layout order $k,j$, so this will lead to the loss of cache hit accuracy and a worse inference speed. Prior work~\cite{10.5555/509058.509096} indicates that a good loop order could bring 20\% to 50\% performance improvement. The memory store pattern also affects the performance. A good store pattern can improve 10\% to 20\% based on architecture platforms. Similar optimization opportunities also exist for CNN~\cite{intel_cnn}. 

\begin{figure}
    \centering
    \includegraphics[width=0.5\linewidth]{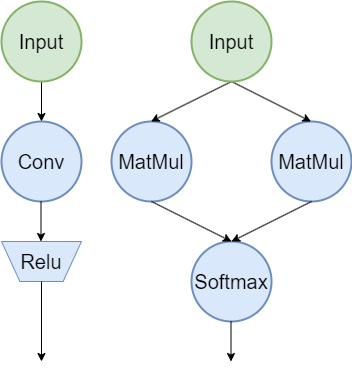}
    \caption{Mockup example for the optimal memory layout challenge: graphs with simple connections (left) can be solved optimally in linear time, while complicated connections (right) and dependencies make the search untractable.}
    % \caption{For the left side model, the dynamic programming algorithm works correctly. For the right side model, because of the bifurcations in this model, dynamic programming cannot get the correct answer.}
    \label{fig:dp}
\end{figure}

While optimizing the memory layout for each individual operator is promising, applying them to a complicated neural network graph can easily become untractable. In the deployment phase, there are numerous heterogeneous operators, each with a few diverse memory layout options. Selecting the optimal layout for each tensor to ultimately achieve global optimality is NP-hard. Early work~\cite{234946} attempted on simple CNNs but failed to address this complication. 
On top of the non-trivial optimization problem, modern transformers and foundation models come with even more sophisticated interweaving structures, with skip connections and arbitrary dependencies between operators throughout the graph. 
To briefly illustrate the challenge using a simple example shown in \figref{fig:dp}, a single branch of operators (left) can be easily optimized~\cite{234946} by searching the best memory layout (\eg using dynamic programming), while the same approach applied on a node with multiple dependencies (right) could result in a layout conflict. The problem manifests into untractable search when the graph becomes bigger and dependencies become more complicated. \sysname aims at a graph-agnostic generic design to address this challenge.

\parab{Peak Memory Footprint and Memory Management.}
The memory space management problem is widely discussed in computer science, especially for the operating system. Researchers proposed a lot of algorithms solving this problem, including First-Fit~\cite{knuth97}, Best-Fit~\cite{knuth97}, Buddy-System~\cite{615762}, SLAB~\cite{10.5555/1267257.1267263}, \etc There is a significant difference between memory allocation problems in operating systems and model reasoning; the operating systems can only predict future allocation requests from users to ``guess'' the best decisions based on the current information they have. On the contrary, model inference is quite predictable. The developers usually freeze the structures and parameters after training to fix the memory allocation requirements of tensors. Therefore, the allocator can find an optimal memory allocation scheme based on global information to achieve the minimum peak memory overhead, which is impossible for the allocation algorithms in the operating systems.

On the topic of static memory allocation algorithms, Gergov proved it as an NP-complete problem in 1999~\cite{10.5555/314500.315082}, so it's impossible to find an optimal solution for this problem in linear time. 
Pisarchyk and Lee introduced that the greedy algorithm is an excellent solution to this problem~\cite{DBLP:journals/corr/abs-2001-03288}. This algorithm is also adopted in TensorFlow-Lite Micro~\cite{DBLP:journals/corr/abs-2010-08678} and other works~\cite{DBLP:journals/corr/abs-1907-01989}. \sysname takes the inspiration and aims at memory management policy coherent with the proposed graph optimization strategies.

\section{\sysname Design}
\label{sec:design}

To break the shackles of rigid inference execution for modern neural networks with complicated graphs, we propose \sysname, a holistic optimization framework to transform the execution blueprint with optimized memory layout and memory management.

% We focus on Ahead-of-Time compilation, memory space management, and the memory layout schedule to improve the inference performance of deep learning models on the CPU.

% \subsection{Dynamic Programming}

% To achieve high-performance code generation, we developed a framework to generate the code we need instead of invoking functions from the high-performance computation libraries. It provides flexibility in extending our algorithm to more tensor memory layout scenarios. 
% For the second problem, we propose a new version of the dynamic programming algorithm to solve the above problems.

\parab{Holistic Graph-Operator Optimization.} Conceptually, \sysname jointly optimizes the memory layout throughout the graph by splitting the whole graph into multiple processable subgraphs and carefully resolves the conflicts among subgraphs. Specifically, \sysname traces the most extended sequence in the computation graph as a first step. It marks every input edge with a distance of 0 and uses a modified variant of the Dijkstra algorithm to walk through the whole graph and record the longest distance to every edge. Then, we traverse all output edges, find the one with the longest distance, and reverse from it to get the most extended sequence. The pseudo code for the longest sequence extraction is shown in \algoref{alg:fls}.

\begin{algorithm}
    \caption{LongestSequenceExtraction}
    \label{alg:fls}
    \begin{algorithmic}
        \STATE \textbf{Function} find-longest-sequence($flow$)
            \STATE $mainflow=\{\}$
            \STATE $queue=\{\}$
            \FOR{$input\in flow$}
                \STATE put the tuple $(input,0)$ in $queue$
            \ENDFOR
            \WHILE{$queue$ isn't empty}
                \STATE pop the first node from $queue$ as $(node,distance)$
                \FOR{$next\in node.successor$}
                    \IF{$next$ is never visited}
                        \STATE put $(next,distance+1)$ in $queue$
                    \ENDIF
                \ENDFOR
            \ENDWHILE
            \STATE find the $node$ with the longest distance
            \STATE reverse the $node$ and get the $sequence$ to the begin node
            \STATE \textbf{return} $sequence$
    \end{algorithmic}
\end{algorithm}

With the longest sequence in the graph, we extract the sequence from the whole graph, cutting all connections so the entire graph is segmented into several independent connected graphs due to the absence of the longest sequence. Then, we extract the longest sequence within each subgraph by running the same algorithm 
% on every connected graph 
recursively until every node is counted in one longest sequence in a subgraph. The recursive extraction is shown in \algoref{alg:split}
% Finally, we get several sequences suitable for dynamic programming.
\begin{algorithm}
    \caption{RecursiveExtraction}
    \label{alg:split}
    \begin{algorithmic}
        \STATE \textbf{Function} split($flow$)
            \STATE $sequnece$ = find-longest-sequence($flow$)
            \STATE remove $sequence$ from $flow$
            \STATE $sequences=\{sequnece\}$
            \STATE create several connected graph $subflows$ on remained nodes
            \FOR{$subflow\in subflows$}
                \STATE $subsequences$=find-longest-sequence($subflow$)
                \FOR{$subsquence\in subsequences$}
                    \STATE put the $subsquence$ in $sequences$
                \ENDFOR
            \ENDFOR
            \STATE \textbf{return} $sequences$
    \end{algorithmic}
\end{algorithm}

% \parab{Dynamic Programming} 
Taking the set of sequences, we run dynamic programming on them to schedule the most efficient memory layout. Specifically, we record the shortest execution time with the specific memory layout to reach every edge. The shortest execution time is computed following \eqnref{eq:dp}, 
\begin{equation}
    t_{v,l} = \min(\sum_{u\in N(v),l'\in L(u)}t_{u,l'}+T_{(u,l'),\cdots,(v,l)})
    \label{eq:dp}
\end{equation}
%%where $t$ is the shortest execution time to reach this edge \hqrev{what edge, explain u, l'}\jjrev{talk about this} and $T$ is the shortest execution time of this kernel \hqrev{same what kernel}.
where $v,l$ is the current edge and layout on scheduling, $u$ is the neighbor of $v$, $l'$ is the possible layout of $u$, $t_{u,l'}$ is the shortest execution time to reach $u$ with layout $l'$, and $T_{(u,l'),\cdots,(v,l)}$ is the execution time of the kernel the specific input and output memory layouts. Then, for the output edge, we select the layout with the shortest execution time and reverse it to the input so we can schedule a memory layout for every edge on the sequence.
The pseudo-code for applying dynamic programming on each sequence is shown in \algoref{alg:dpos}

\begin{algorithm}
    \caption{DPOnSequence}
    \label{alg:dpos}
    \begin{algorithmic}
        \STATE \textbf{Extern Function} eval($node$, $layouts$) \COMMENT{given the node and input and output layouts, the algorithm will return its time cost in the real world}
        \STATE \textbf{Function} dp-on-sequence($sequence$)
            \STATE $table=\{\}$
            \FOR{$edge\in sequence$}
                \FOR{$layout\in$ all possible layouts of $edge$}
                    \STATE $mintimecost=\min\sum(\sum\min timecosts\text{ of previous edges}+\text{eval}(node,layouts))$
                \ENDFOR
                \STATE find the $layout$ corresponding to the min time cost
                \STATE save $(edge,layout)$ in $table$
            \ENDFOR
            \STATE \textbf{return} $table$
    \end{algorithmic}
\end{algorithm}

% \parab{Merge.} Now, 
Finally, with each sequence, we have an optimized blueprint for its memory layout and execution pattern. However, an edge could exist in several sequences, and we may have scheduled it with different layouts during the dynamic programming, since the optimal strategy is confined within each sequence. To resolve memory layout conflicts, we cautiously analyze all dependencies, and adopt a majority voting scheme to select the layout option that can benefit the most sequences. The conflicting sequence would have to compromise; re-optimizing with the given constraints of the resolved layout options.  We merge all sequences back to the whole graph with the memory layout schedule information for every edge.

\parab{Loop Reordering.} For certain operators such as matrix multiplication, we also introduce a loop reorder mechanism to optimize their execution efficiency. As discussed above, the order of loops also affects the pattern of memory access, thus affecting execution speed. For these operators, we will try various loop orders under the influence of various input and output memory layouts, and save the fastest among them as the choice for dynamic programming and final implementation.

% Greedy-by-Size Algorithm
\begin{algorithm}
    \caption{Greedy-by-Size}
    \label{alg:greedy}
    \begin{algorithmic}
        \STATE \textbf{Function} greedy-by-size($tensors$)
            \STATE sort $tensors$ by order of size descending
            \WHILE{$tensors$ is not empty}
                \STATE $tensor$ = pop front $tensors$
                \STATE delete the available space of the tensor whose lifetime overlaps with $tensor$'s
                \FOR{$slot\in slots$}
                    \IF{$slot.size>=tensor.size$}
                        \STATE allocate $slot$ to $tensor$
                        \STATE \textbf{break}
                    \ENDIF
                \ENDFOR
                \IF{$tensor$ isn't allocated}
                    \STATE increase and allocate the last free available block
                \ENDIF
            \ENDWHILE
    \end{algorithmic}
\end{algorithm}

\begin{figure}
    \centering
    \includegraphics[width=\linewidth]{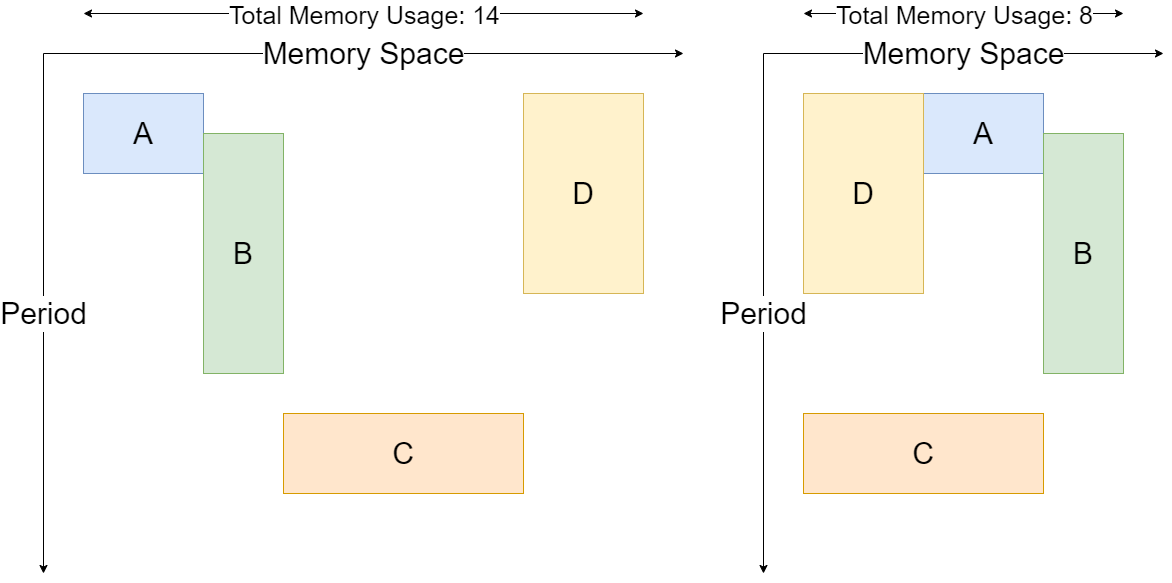}
    \caption{Memory Allocation Strategy. The left is the naive version, and the right is the dynamic programming version.}
    \label{fig:memory}
\end{figure}

\parab{Memory Allocation and Management.}
One straightforward implementation of a memory allocator in the model inference is to use the dynamic allocator provided by the operating system, like "ptmalloc", "jemalloc" and so on. Because we know the memory requirements in the inference process, we can allocate them in advance instead of in the runtime to save time. So, as shown in \figref{fig:memory}, we can first allocate a sequential memory space and a new slice to every tensor. However, we can see a potential optimization. Some tensors do not overlap in their lifetime, so they can share the same memory space at different periods.

\sysname leverages a \textit{Greedy-by-Size} strategy to solve this problem as shown in \figref{fig:memory}. At the beginning, we sort all tensors by size in a queue. Then, we take the tensor with the most significant size from the queue every time, try to find the most suitable slot in the current memory space, and allocate it for the tensor. If no slot can hold the tensor, we will increase our current memory space until the free slot at the end is large enough to hold the tensor. \figref{fig:memory} shows that the peak memory usage is less than the naive allocation. More algorithm details are in the pseudo-code \algoref{alg:greedy}.

\section{\sysname Implementation}

\begin{figure}
    \centering
    \includegraphics[width=0.8\linewidth]{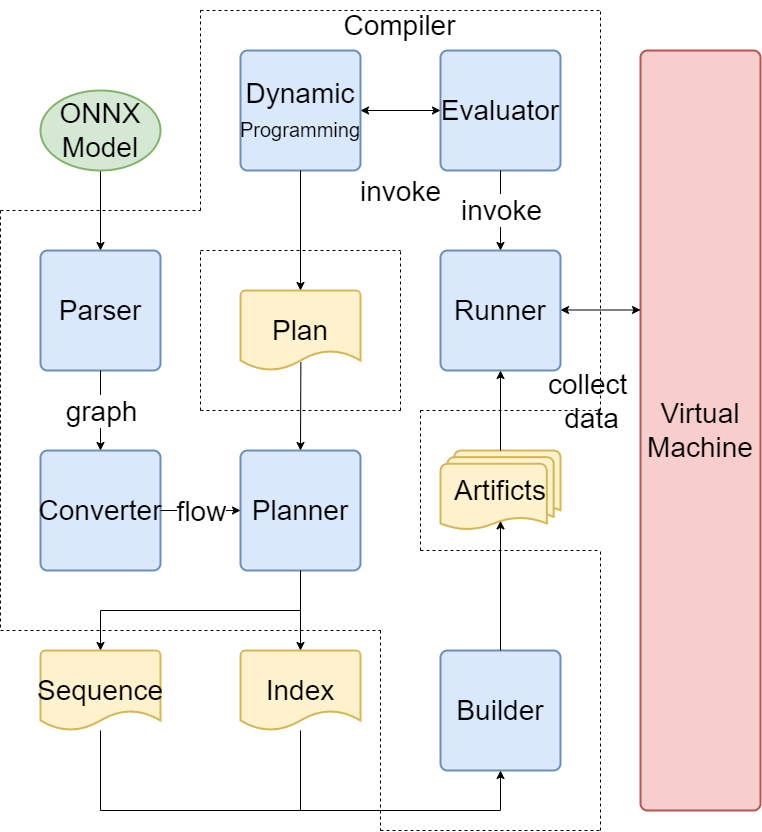}
    \caption{\sysname Workflow}
    \label{fig:workflow}
\end{figure}

{\sysname} is built based on MLIR, which supports developing general-purpose deep learning compilers. The \sysname architecture is split into two main components: a compiler and a virtual machine. We wrote approximately 30K lines of C++ codes to implement it, including an MLIR-base compiler, a virtual machine to execute the generated code and collect data from the real world, reimplementing widely-used kernels, and \sysname's core optimizations framework. To achieve high-performance code generation, we developed a framework to generate the code we need instead of invoking functions from the high-performance computation libraries. It provides flexibility in extending our algorithm to more tensor memory layout scenarios. 
% \subsection{\sysname Architecture and Workflow}
\figref{fig:workflow} shows the workflow of \sysname. 

\parab{Compiler.}
The compiler chooses ONNX as the front-end format. For a valid ONNX model, it converts the model into built-in formats. In this stage, some operator fusions and memory usage optimization happen. Then, it converts the data structure from a graph into a linear sequence, so it's convenient for the later stage to transform it into MLIR codes. Also, the memory layout reschedule happens in this stage within the compiler.

Then, the compiler compiles the linear sequence into MLIR codes. We implemented tens of widely used operators in ONNX. We built our compilers mainly based on "Linalg," "Affine," "Memref," \etc, instead of creating a new dialect. We select "Memref" instead of "Tensor" because we can control more details of memory layouts, consume fewer memory spaces, and have better memory accessing speed. We utilize the passes provided by MLIR to lower the generated codes to LLVM IR so they are compatible with popular frameworks, lowering the barrier for user adoption.

\parab{Virtual Machine.}
To have better evaluation results of our code generation, we developed a virtual machine to run the code dynamically. The virtual machine is built on the Just-in-Time(JIT) Execution Engine provided by MLIR, which could run the code produced by the compiler. We also provide an interface so that the users can invoke functions compiled by JIT easily. There are two main functions of the virtual machine. Firstly, it provides a dynamic way to invoke the code so users can use it more flexibly. After that, the virtual machine is also an essential component in the optimization stage, which collects the actual performance data and helps the compiler find more opportunities.

\begin{figure*}
    \centering
    \begin{minipage}{0.66\columnwidth}
        \centering
        \includegraphics[width=\columnwidth]{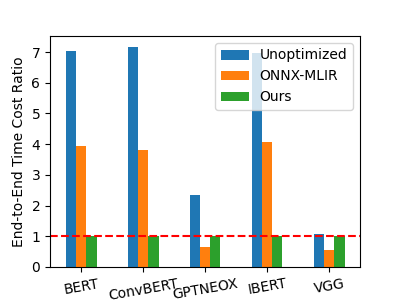}
        \caption{Normalized Latency on AMD}
        \label{fig:amd-end2end}
    \end{minipage}
    \hfill
    \begin{minipage}{0.66\columnwidth}
        \centering
        \includegraphics[width=\columnwidth]{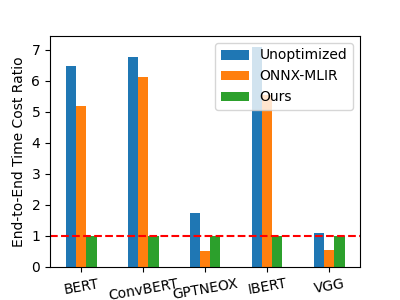}
        \caption{Normalized Latency on Intel}
        \label{fig:intel-end2end}
    \end{minipage}
    \hfill
    \begin{minipage}{0.66\columnwidth}
        \centering
        \includegraphics[width=\columnwidth]{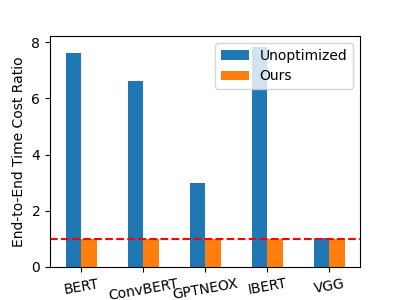}
        \caption{Normalized Latency on Apple}
        \label{fig:apple-end2end}
    \end{minipage}
\end{figure*}

\section{Evaluation}
\label{sec:eval}

We evaluate \sysname on three platforms: Intel i7-13700 with 32GB RAM, AMD 5955WX with 128GB RAM, and Apple M1 Pro with 8GB RAM. All of them work on Ubuntu 22.04, while Apple has the support of Docker. The hardware platform details are summarized in \tabref{tab:hw_stats}. We adapted BERT~\cite{DBLP:journals/corr/abs-1810-04805}, ConvBERT~\cite{DBLP:journals/corr/abs-2008-02496}, GPT-NEOX~\cite{gpt-neox-library}, I-BERT~\cite{kim2021bert} and VGG~\cite{7486599} with \sysname, and use them to evaluate the performance of \sysname on different aspects.

% \begin{table}
%     \centering
%     \begin{tabular}{c|c|c|c}
%         % \hline
%         Vendor & Intel & AMD & Apple \\ \hline
%         Arch & x86-64 & x86-64 & aarch64 \\ 
%         % \hline
%         Model & i7-13700 & 5955WX & M1 Pro \\ 
%         % \hline
%         Cores & 24 & 32 & 16 \\ 
%         % \hline
%         OS & Ubuntu & Ubuntu & Ubuntu \\ 
%         % \hline
%         RAM(GB) & 32 & 128 & 8 \\ 
%         % \hline
%     \end{tabular}
%     \caption{Evaluation Hardware Platforms}
%     \label{tab:hw_stats}
% \end{table}

\begin{table}
    \centering
    \begin{tabular}{c|c|c|c}
        % \hline
        Vendor & Intel & AMD & Apple \\ \hline
        Arch & x86-64 & x86-64 & aarch64 \\ 
        % \hline
        Model & i7-13700 & 5955WX & M1 Pro \\ 
        % \hline
        Cores & 24 & 32 & 16 \\ 
        % \hline
        OS & Ubuntu & Ubuntu & Ubuntu \\ 
        % \hline
        RAM(GB) & 32 & 128 & 8 \\ 
        % \hline
    \end{tabular}
    \caption{Evaluation Hardware Platforms}
    \label{tab:hw_stats}
\end{table}

We evaluate \sysname on the models mentioned before across the above platforms. The models we selected include conventional vision models and the newly emerging LLMs. We hope to include different types of models to verify how \sysname performs in various scenarios. For baseline, we select ONNX-MLIR~\cite{DBLP:journals/corr/abs-2008-08272}, which follows a similar technical road map as ours, so that we can better highlight the enhancement of our framework compared to theirs. Due to the lack of support for the docker image of ONNX-MLIR on the aarch64 Linux system, ONNX-MLIR is only evaluated and compared on x86-64 platforms.

% \subsection{Inference Efficiency Evaluation}

% \subsection{End-to-End Performance}

\begin{table}
    \centering
    \small
    \begin{tabular}{c|c|c|c|c|c}
        % \hline
         & BERT & GPT & IBERT & Conv & VGG \\
         &  & NEOX &  & BERT & \\
         & (sec) & (msec) & (sec) & (sec) & (sec) \\ \hline
        \multicolumn{6}{c}{AMD} \\ \hline
        Unoptimized & 34.2 & 21 & 32.3 & 34.2 & 19.6 \\ \hline
        Optimized & 4.8 & 9 & 4.6 & 4.8 & 19.0 \\ \hline
        ONNX-MLIR & 19.1 & 6 & 18.9 & 18.2 & 10.1 \\ \hline
        \multicolumn{6}{c}{Intel} \\ \hline
        Unoptimized & 22.7 & 14 & 21.9 & 26.0 & 10.8 \\ \hline
        Optimized & 3.5 & 8 & 3.1 & 3.8 & 10.0 \\ \hline
        ONNX-MLIR & 18.3 & 4 & 17.1 & 23.5 & 5.5 \\ \hline
        \multicolumn{6}{c}{Apple} \\ \hline
        Unoptimized & 42.2 & 60 & 40.6 & 37.5 & 55.4 \\ \hline
        Optimized & 5.5 & 20 & 5.2 & 5.7 & 54.1 \\ 
        % \hline
    \end{tabular}
    \caption{End-to-end Inference Latency}
    \label{tab:inference-timecost}
\end{table}

\subsection{End-to-end Performance}

The end-to-end latency result is shown \figref{fig:amd-end2end}, \figref{fig:intel-end2end}, and \figref{fig:apple-end2end}. To highlight \sysname's enhancement, we show the other comparison latency normalized by \sysname latency while listing the absolute latency numbers in \tabref{tab:inference-timecost} for reference. We compare the time cost of \sysname (optimized) with its unoptimized version and ONNX-MLIR. The unoptimized version performs the worst and has the longest inference time on all models and platforms. \sysname significantly outperforms ONNX-MLIR on BERT, ConvBERT, and I-BERT, but slightly lags behind on GPT-NEOX and VGG. This is related to the matrix multiplication feature of their shapes. We will discuss this issue more in the next section.

\subsection{Ablation Study}

To better analyze the benefits brought to different operators due to different optimization methods, we provide an interface for \sysname to generate intermediate results for our analysis. In terms of our inference speed optimization, we mainly considered two approaches. First, for all operator types, we consider optimizing their tensor memory layout through dynamic programming to improve the execution speed of the operators. Secondly, we consider reordering loops for special operators such as matrix multiplication to make the memory access mode more cache-friendly, thereby improving the inference speed. We separate these two aspects of information collected by the compiler during the compilation, process and export them separately, and then analyze them to obtain the results. Here, we conduct experiments on the Intel platform, and the results are shown in \figref{fig:op_res} and \figref{fig:kernel_res} (see Appendix). The first method is shown in orange (\sysname without reordering), and the second is shown in green (\sysname with reordering)\footnote{A few cases/bars are missing due to hardware limitations forbidding us from getting the evaluation results.}

% \begin{figure}
%     \centering
%     \includegraphics[width=\linewidth]{fig/intel-bert-bar.png}
%     \caption{BERT Optimization Ratio}
%     \label{fig:bert-bar}
% \end{figure}
% \begin{figure}
%     \centering
%     \includegraphics[width=\linewidth]{fig/intel-convbert-bar.png}
%     \caption{ConvBERT Optimization Ratio}
%     \label{fig:convbert-bar}
% \end{figure}
% \begin{figure}
%     \centering
%     \includegraphics[width=\linewidth]{fig/intel-gptneox-bar.png}
%     \caption{GPT-NEOX Optimization Ratio}
%     \label{fig:gptneox-bar}
% \end{figure}
% \begin{figure}
%     \centering
%     \includegraphics[width=\linewidth]{fig/intel-ibert-bar.png}
%     \caption{IBERT Optimization Ratio}
%     \label{fig:ibert-bar}
% \end{figure}

\begin{figure*}
    \centering
    \subfigure[MatMul]{
    \includegraphics[width=0.65\columnwidth]{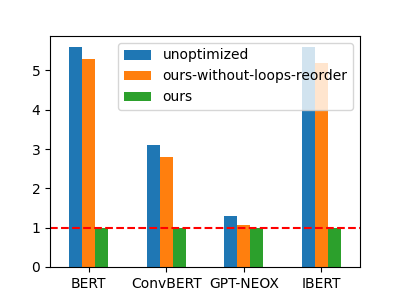}
    }
    \centering
    \subfigure[GEMM]{
    \includegraphics[width=0.65\columnwidth]{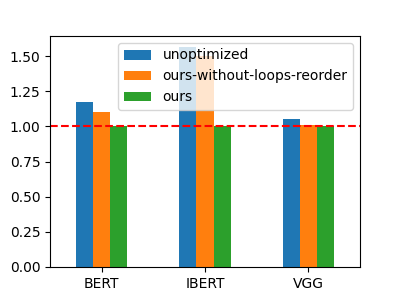}}
    \centering
    \subfigure[Conv]{
    \includegraphics[width=0.65\columnwidth]{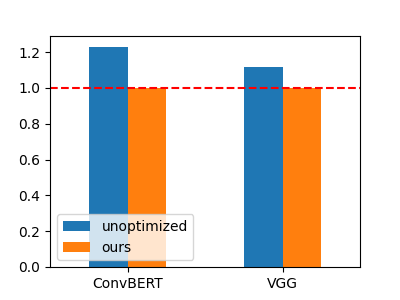}}
    \caption{
    % \jjrev{I select three computation-intensive operators to show them there and move the previous one to the appendix. Does it look good? TODO: give it a name}
    Computation-Intensive Operators Latency Reduction on Different Models Normalized by Unoptimized
    % \hang{Put caption, double check the text that refers to the previous figures, correct them to use the new label.}
    % \hang{Can we move all of these five figures into appendix, instead, shall we pick three common operators and combine them into three graphs (eg. MatMul, GEMM, and something), each having all models' results for a particular operator? ok to not have all models' results for the third operator if not all models have it }
    }
    \label{fig:op_res}
\end{figure*}

We recorded the time cost proportions of the six operators with the most enormous time consumption. In the figure, we set the unoptimized version's time consumption as the unit's baseline and calculate the optimized time cost on this basis. For ordinary operators, their optimization is brought about by dynamic programming. For operators related to matrix multiplication, further optimizations come from rearranging loops. \figref{fig:op_res} shows that the ordinary operator contains two bars representing the unoptimized version and the dynamic programming optimized version, respectively. MatMul and Gemm also have an additional bar representing the optimization after enabling axis reordering.

\begin{figure}
    \begin{subfigure}
        \centering
        \includegraphics[width=\linewidth]{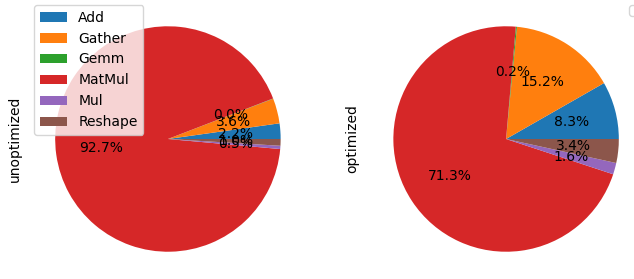}
        \caption{Time Cost Ratio of Different Kernels in BERT 
        % \hang{Text is all cluttered together, is there away to separate them to make it clear?}\jinjie{I haven't found ways to separate them, but I use legend instead here. How does it look?}
        }
        \label{fig:bert-pie}
    \end{subfigure}
    \begin{subfigure}
        \centering
        \includegraphics[width=\linewidth]{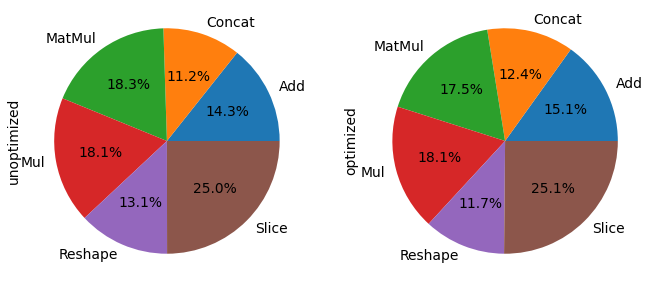}
        \caption{Time Cost Ratio of Different Kernels in GPT-NEOX}
        \label{fig:gptneox-pie}
    \end{subfigure}
\end{figure}

At the same time, we noticed that the benefits brought by loop reorder to the MatMul operator are huge. It provides gains of 76.6\%, 67.7\%, and 75.0\% in BERT, ConvBERT, and I-BERT, respectively. Considering that the main computational cost in the Transformers model is matrix multiplication~\cite{DBLP:journals/corr/VaswaniSPUJGKP17,DBLP:journals/corr/abs-2009-06732}, the overall benefit brought by loop reorder is considerable. At the same time, we noticed that for the Gemm operator. However, the loop reorder has also brought significant improvements; it has brought about 8\%, 10.0\%, and 32\% acceleration in BERT, ConvBERT, and I-BERT respectively, but it is smaller compared to MatMul. We guess this is because GEMM includes bias. In our implementation, this part does not participate in optimizing loop reorder, so it may destroy the cache-friendly memory access pattern. However, we also noticed that the situation is slightly different for GPT-NEOX. The approximately 18.3\% benefit brought by dynamic programming is greater than the approximately 4.4\% benefit brought by loop reorder. By analyzing the network structure, we propose two possible reasons. First, the parameter amount of the matrix multiplication operator in GPT-NEOX is small, so the memory overhead of matrix multiplication itself is small, and the memory access benefit brought by the reorder loop is not apparent. Secondly, the matrix multiplication shape in GPT-NEOX is more rectangular than square, which may also affect the benefits of the final loop reorder. 
It is interesting that this exactly echoes the phenomenon in the previous section. The matrix multiplication characteristics of GPT-NEOX resulted in \sysname's optimization results not being noticeable enough. However, the most crucial matrix multiplication overhead has been significantly optimized for BERT and I-BERT, so \sysname surpassed ONNX-MLIR. However, the small matrix multiplication optimization is insufficient for GPT-NEOX to perform better than ONNX-MLIR. Another evidence is shown in the \figref{fig:bert-pie} and \figref{fig:gptneox-pie}. The overhead proportion of matrix multiplication dropped rapidly in BERT, but this did not happen in GPT-NEOX. 

ConvBERT introduces another computationally intensive operator: the convolution operation. It can be seen that the convolution operator has also achieved certain optimization results under \sysname. As for VGG, its improvement is not very obvious. The reason is that there are not a large number of matrix multiplication operators.
% , and the shape of the matrix multiplication operator is also not help. 
Therefore, \sysname lacks enough optimization opportunities to improve its performance. 
% As mentioned above, \sysname relies on the improvement of the matrix multiplication operator to achieve the performance over ONNX-MLIR, and when this opportunity is missing, the optimization effect is relatively limited.}
%
For more information on different operators' performances on the different models, please refer to \figref{fig:kernel_res} in the appendix.

% \begin{figure}
%     \begin{subfigure}
%         \centering
%         \includegraphics[width=\linewidth]{fig/intel-bert-pie.png}
%         \caption{Time Cost Ratio of Different Kernels in BERT \hang{Text is all cluttered together, is there away to separate them to make it clear?}}
%         \label{fig:bert-pie}
%     \end{subfigure}
%     \begin{subfigure}
%         \centering
%         \includegraphics[width=\linewidth]{fig/intel-gptneox-pie.png}
%         \caption{Time Cost Ratio of Different Kernels in GPT-NEOX}
%         \label{fig:gptneox-pie}
%     \end{subfigure}
% \end{figure}

\subsection{Cost of Conflict Resolution}

When we introduce merging different sequences, the same edge may be scheduled into different layouts in various sequences. We need a way to resolve this conflict. The method we use is to select the design that is scheduled the most times from the conflicting edge candidate layouts. If multiple layouts share the maximum number of times, one is randomly chosen among them. Although this resolves the conflict, it obviously also destroys the optimality of the algorithm. We need to evaluate how much loss this solution will bring to the final result of the algorithm.

\begin{figure}
    \centering
    \subfigure[]{
        \centering
        \includegraphics[width=0.4\linewidth]{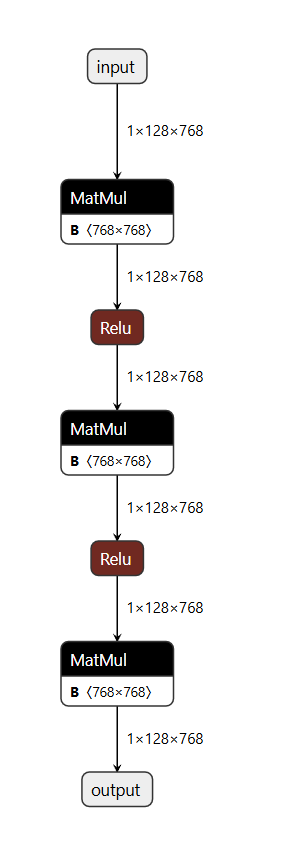}
    }
    \subfigure[]{
        \centering
        \raisebox{0.4\height}{\includegraphics[width=0.5\linewidth]{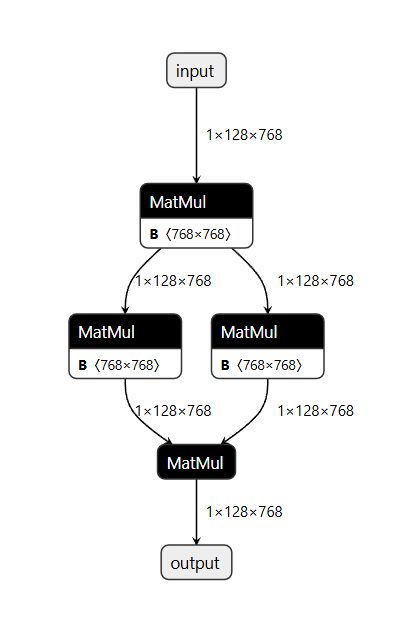}}

    }
    \vfill
    \caption{Toy example models for upper-bound comparison. 
    % \hang{Why is (b) so blurry?} \jinjie{Does this look better? I adjusted their widths, and they don't share the same weight now. I guess it's because their width differs, and LaTeX scales them to match the same width.}
    }
    \label{fig:mlp}
\end{figure}

To verify this problem, we designed two smaller models, used the Brute Force algorithm to find its optimal plan, and compared the results with the plan found by \sysname. These two models represent linear neural networks and branched neural networks, respectively. We want to verify that \sysname's conflict resolution works well enough in both cases. The model structures are shown in \figref{fig:mlp}; the result is in \tabref{tab:layout_conflict}. We found that compared with the global optimal situation, the layout of \sysname schedule will only lose about 17\% and 9\% of the speed in the two models, respectively. This loss rate is acceptable, so we can consider this method effective.

\begin{table}
    \centering
    \begin{tabular}{c|c|c}
         & Optimal Schedule & \sysname Schedule \\ \hline
        Model(a) & 130.5 ms & 156.5 ms \\ \hline
        Model(b) & 138.9 ms & 152.0 ms \\
    \end{tabular}
    \caption{Comparison to optimal schedule using brute-force search. \sysname is very close to upper-bound schedule performance while keeping the computing overhead linear.}
    \label{tab:layout_conflict}
\end{table}

\subsection{Memory Management Evaluation}

To evaluate how our memory management works in the real world, we choose three popular frameworks as the baselines: ONNXRuntime~\cite{onnxruntime}, PyTorch~\cite{DBLP:journals/corr/abs-1912-01703} and Apache-TVM~\cite{DBLP:journals/corr/abs-1802-04799}. We count the resident set size as the basis for evaluating the program's memory overhead. \sysname and TVM are compilers, and we ignore the memory overhead during the compilation and only count in the runtime memory cost.

\figref{fig:memory} and table \tabref{tab:memory-overhead} show the result on AMD. \sysname performs better than ONNXRuntime and PyTorch on all five models and performs better than TVM on BERT, ConvBERT, and GPT-NEXO. These results show that \sysname works well in minimizing peak memory usage. \sysname performs memory planning based on global information and uses a greedy algorithm to find the optimal solution. Compared with other frameworks based on the operating system memory allocator, it has more decision information and can make better memory allocation plans.

% \begin{table}
%     \centering
%     \begin{tabular}{c|c|c|c|c|c}
%         % \hline
%          & BERT & ConvBERT & GPTNEOX & IBERT & VGG \\
%          & (MB) & (MB) & (MB) & (MB) & (MB) \\ \hline
%         PyTorch & 2615.4 & & 535.7 & 2916.3 & 6113.6 \\ 
%         % \hline
%         ONNXRuntime & 835.7 & 790.8 & 59.1 & 915.1 & 1131.3 \\ 
%         % \hline
%         TVM & 1426.4 & 1381.2 & 144.6 & 142.1 & \\ 
%         \hline
%         \sysname & \textbf{346.6} & \textbf{329.9} & \textbf{3.1} & \textbf{348.0} & \textbf{582.7} \\ 
%         % \hline
%     \end{tabular}
%     \caption{The Memory Overhead\jjrev{PyTorch reports error on ConvBERT and TVM reports error on VGG. Looks like their bugs. How should we explain the reason?} \hang{Put a '*' in the results and add a footnote to explain the bugs}
%     \hang{When no space, use spreadsheet and import them as pdf figure}}
%     \label{tab:memory-overhead}
% \end{table}

\begin{table*}
    \centering
    \begin{tabular}{c|c|c|c|c|c}
        % \hline
         & BERT(MB) & ConvBERT(MB) & GPTNEOX(MB) & IBERT(MB) & VGG(MB) \\ \hline
        PyTorch & 2615.4 & - & 535.7 & 2916.3 & 6113.6 \\ 
        % \hline
        ONNXRuntime & 835.7 & 790.8 & 59.1 & 915.1 & 1131.3 \\ 
        % \hline
        TVM & 1426.4 & 1381.2 & 144.6 & 142.1 & - \\ 
        \hline
        \sysname & \textbf{346.6} & \textbf{329.9} & \textbf{3.1} & \textbf{348.0} & \textbf{582.7} \\ 
        % \hline
    \end{tabular}
    \caption{Memory Overhead Comparison\footnotemark}
    \label{tab:memory-overhead}
\end{table*}
\footnotetext{onnx2torch reports an error on ConvBERT due to it doesn't support the Pad operator, and TVM fails on compiling VGG, so we skip them when measuring the memory overhead.}

\begin{figure}
    \centering
    \includegraphics[width=0.95\linewidth]{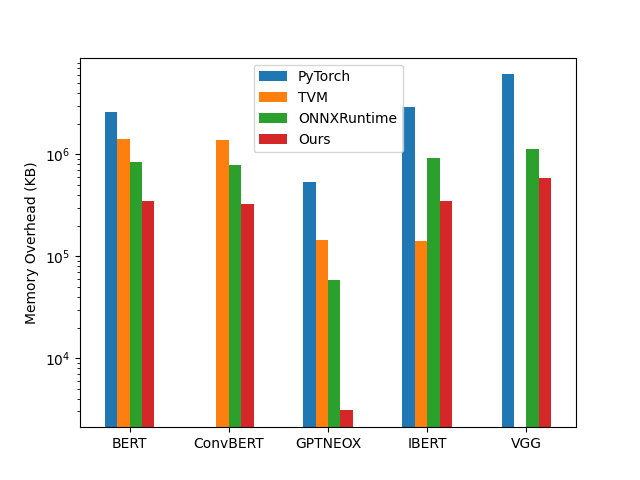}
    \caption{Memory Overhead Comparison} 
    % \hang{This is not how to plot log scale image. use $ax.set\_yscale('log')$}}
    % \hang{Any story we can tell why we choose these five models?} \jinjie{Please refer to the second paragraph in this section.}
    \label{fig:memory-overhead}
\end{figure}

\sysname and TVM are both compiled, so we compare their allocations on some models, and the result is shown in the table \tabref{tab:allocations}. We can see the allocation times of \sysname is almost negligible compared to TVM. This is due to \sysname taking the strategy of static allocation. The user only needs to allocate a big block of buffer, and the compiler divides and assigns them to different tensors during the compilation instead of runtime so that the model inference can save time.

\begin{table}
    \centering
    \begin{tabular}{c|c|c|c}
         & BERT & GPT-NEOX & IBERT \\ \hline
        TVM & 182292 & 158517 & 148103 \\ \hline
        \sysname & 7 & 14 & 7 \\ 
    \end{tabular}
    \caption{Number of Memory Allocations Invocations}
    \label{tab:allocations}
\end{table}

\section{Related Work}
\label{sec:related}

\parab{High-Performance Library.} Many hand-tuned libraries are very popular to support highly efficient model inference. BLAS, with many following works like OpenBLAS~\cite{6413635,6877458}, focuses on optimizing linear algebra kernels. ATLAS~\cite{CLINTWHALEY20013} tries to import auto-tuning in the kernel design for better performance. With the rise of AI, a lot of developers and vendors developed libraries for AI inference, like OpenVINO~\cite{openvino}, OneMKL~\cite{onemkl} for Intel, CMSIS-NN~\cite{DBLP:journals/corr/abs-1801-06601} for Arm, TensorRT~\cite{tensorrt} for Nvidia GPU, and so on. For low-precision inference, Meta developed fbgemm~\cite{DBLP:journals/corr/abs-2101-05615}, qnnpack~\cite{qnnpack} for different architectures, while XNNPACK~\cite{xnnpack} is developed by Google for a similar purpose.

\parab{ML Compiler.} There are a lot of ML compilers growing these years. Halide~\cite{10.1145/2499370.2462176} tries to generate kernel libraries automatically instead of manual tune. MLIR~\cite{DBLP:journals/corr/abs-2002-11054, mlir} provides an infrastructure required for deep learning compiler development, and there are a lot of other projects built on it, including IREE ~\cite{The_IREE_Authors_IREE_2019}, ONNX-MLIR~\cite{DBLP:journals/corr/abs-2008-08272}. XLA~\cite{50530} is a deep learning compiler developed by Google to accelerate some popular frameworks, including Tensorflow~\cite{tensorflow2015-whitepaper}, JAX~\cite{jax2018github} etc. Glow~\cite{DBLP:journals/corr/abs-1805-00907} is a sub-project of PyTorch~\cite{DBLP:journals/corr/abs-1912-01703}. TVM~\cite{DBLP:journals/corr/abs-1802-04799}, Ansor~\cite{258858}, etc, work on optimizing kernels with learning-based ways for types of back-end targets. There are works focusing on optimization for specific models on CPU, including DeepCPU~\cite{216077} and NeoCPU~\cite{234946}. Rammer~\cite{258921} provides a compiler to utilize the parallelism of operators for acceleration. There is also some research on efficient inference for edge devices, like TensorFlow Lite Micro~\cite{MLSYS2021_6c44dc73}, vMCU~\cite{zheng2024vmcucoordinatedmemorymanagement}, MCUNet~\cite{DBLP:journals/corr/abs-2110-15352} and so on.

\section{Conclusion}

In this work, we introduce \sysname, a generic runtime memory management and optimization framework that can flexibly transform the model execution blueprint to achieve faster and more memory-efficient inference. \sysname is a model/graph agnostic framework, jointly optimizes throughout the whole graph to compute memory layout schedule in linear time. It also carefully manages the peak memory usage to shrink the memory footprint. Evaluations across different platforms show that \sysname can consistently reduce the end-to-end inference latency by up to 25.38\% for popular language models and reduce peak memory usage by up to 41.47\%, compared to state-of-the-art approaches. \sysname is of $\sim$30K line of codes, built for general-purpose usage, and will be released as an open-source inference runtime optimization framework to the community.

\balance
% \newpage
\bibliography{references}

\bibliographystyle{mlsys2025}

\newpage
\appendix

\section{Latency Improvement per Operator}

\begin{figure*}[b]
    \centering
    \subfigure[BERT]{
    \includegraphics[width=0.65\columnwidth]{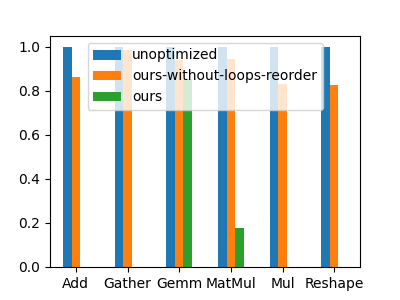}
    }
    \centering
    \subfigure[ConvBERT]{
    \includegraphics[width=0.65\columnwidth]{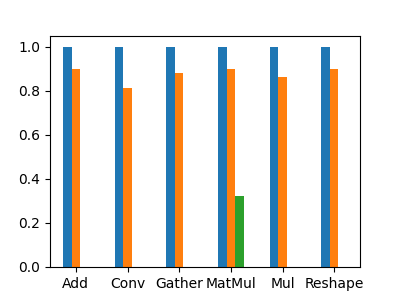}}
    \centering
    \subfigure[GPT-NEOX]{
    \includegraphics[width=0.65\columnwidth]{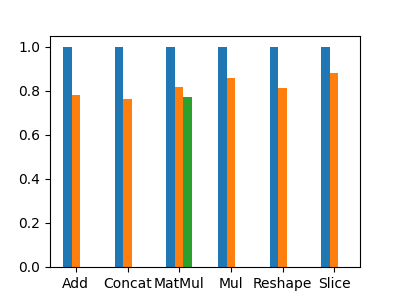}}
    \centering
    \subfigure[IBERT]{
    \includegraphics[width=0.65\columnwidth]{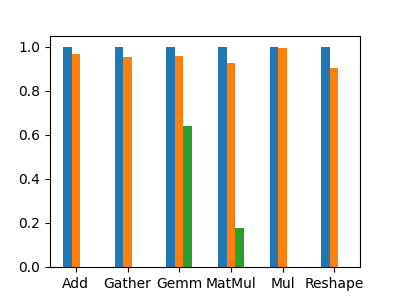}}
    \centering
    \subfigure[VGG]{
    \includegraphics[width=0.65\columnwidth]{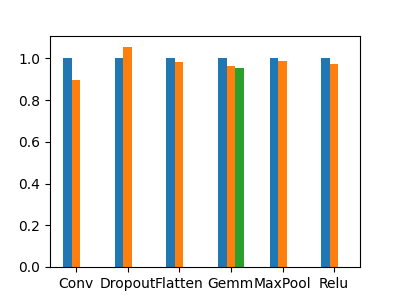}}
    \caption{Normalized Latency Reduction per Operator}
    \label{fig:kernel_res}
\end{figure*}

\end{document}